\documentclass[10pt,twocolumn,letterpaper]{article}

\usepackage[accsupp]{axessibility}

\usepackage{iccv}
\usepackage{times}
\usepackage{epsfig}
\usepackage{graphicx}
\usepackage{amsmath}
\usepackage{amssymb}
\usepackage{bbm}
\usepackage{todonotes}

\usepackage{booktabs}
\usepackage{bm}
\usepackage{multirow}
\usepackage{multicol}

\usepackage[ruled,vlined]{algorithm2e}

\usepackage{setspace}

\usepackage{subfig}

\newcommand{\vect}[1]{{#1}}

\newcommand{\set}[1]{\mathcal{#1}}

\usepackage[breaklinks=true,bookmarks=false,]{hyperref}

\hypersetup{
    colorlinks=true,
}
\iccvfinalcopy 

\def\iccvPaperID{4446}

\ificcvfinal\fi

\makeatletter
\def\thanks#1{\protected@xdef\@thanks{\@thanks
        \protect\footnotetext{#1}}}
\makeatother

\begin{document}

\title{On the Robustness of Open-World Test-Time Training: Self-Training with Dynamic Prototype Expansion}

\author{\and \and Yushu Li $^1$ \and Xun Xu $^{2}$ \and Yongyi Su $^1$ \and Kui Jia $^{1}$ \and \and $^1$ South China University of Technology \and $^2$ Institute for Infocomm Research (I2R), Agency for Science, Technology and Research (A*STAR) \and {\tt\small \{eeyushuli, eesuyongyi\}@mail.scut.edu.cn, \{alex.xun.xu, kuijia\}@gmail.com}}

\maketitle
\ificcvfinal\thispagestyle{empty}\fi

\begin{abstract}
    Generalizing deep learning models to unknown target domain distribution with low latency has motivated research into test-time training/adaptation~(TTT/TTA). Existing approaches often focus on improving test-time training performance under well-curated target domain data. As figured out in this work, many state-of-the-art methods fail to maintain the performance when the target domain is contaminated with strong out-of-distribution~(OOD) data, a.k.a. open-world test-time training~(OWTTT). 
    The failure is mainly due to the inability to distinguish strong OOD samples from regular weak OOD samples. To improve the robustness of OWTTT we first develop an adaptive strong OOD pruning which improves the efficacy of the self-training TTT method. We further propose a way to dynamically expand the prototypes to represent strong OOD samples for an improved weak/strong OOD data separation. Finally, we regularize self-training with distribution alignment and the combination yields the state-of-the-art performance on 5 OWTTT benchmarks. The code is available at \url{https://github.com/Yushu-Li/OWTTT}.

\end{abstract}

\vspace{-0.1cm}
\section{Introduction}

The distribution gap between training and testing data poses great challenges to the generalization of modern deep learning methods~\cite{quinonero2008dataset,ben2010theory}. To improve model's generalization to testing data which may feature a different data distribution from the training data, domain adaptation has been extensively studied~\cite{wang2018deep} to learn domain invariant features. Nevertheless, the existing unsupervised domain adaptation paradigm requires simultaneous access to both source and target domain data with an off-line training stage~\cite{YaroslavGanin2015UnsupervisedDA,tang2020discriminative}. In a realistic scenario, access to target domain data may not become available until the inference stage, and instant prediction on testing data is required without further ado. Therefore, these requirements give rise to the emergence of a new paradigm of adaptation at test time, a.k.a. test-time training/adaptation~(TTT/TTA)~\cite{YuSun2019TestTimeTW,wang2020tent}.

\begin{figure}[!tb]
    \centering
    \includegraphics[width=0.9\linewidth]{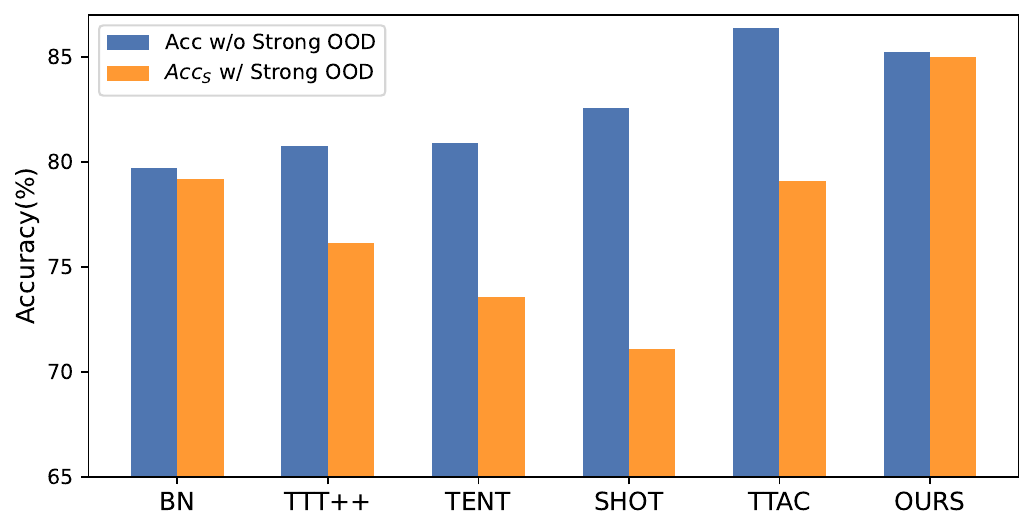}
    \vspace{-0.2 cm}
    \caption{Existing test-time training methods suffer substantially when target domain data is contaminated with strong OOD samples. We illustrate the results by test-time training on CIFAR10-C contaminated by SVHN as strong OOD samples, with reference to values extracted from Tab~\ref{tab:CIFAR10-C}.}
    \label{fig:performance_drop}
    \vspace{-0.5 cm}
\end{figure}

The success of TTT has been demonstrated on many synthesized corrupted target domain data~\cite{hendrycks2019benchmarking}, manually selected hard samples~\cite{recht2018cifar} and adversarial samples~\cite{croce2022evaluating}. Nevertheless, the boundary of existing TTT methods' capability is yet to be fully explored. In particular, to enable TTT in the open-world, focus has been shifted to investigating open-world scenarios where TTT methods could fail. Among these scenarios, when testing data is not drawn in an i.i.d. fashion, TTT methods may be biased towards the continually changing distribution~\cite{wang2022continual,gongnote2022}. When the target domain consists of testing data drawn from both source and target distributions, \cite{niu2022efficient} developed a non-forgetting training paradigm to avoid failure on source domain samples. When TTT must be updated with small batch size, shifting distribution, and class imbalanced testing data, \cite{niu2023towards} proposed to swap out the batch normalization and remove unreliable pseudo labels for improved robustness.

Despite many efforts into developing stable and robust TTT methods under a more realistic open-world environment, in this work, we delve into an overlooked, but very commonly seen open-world scenario where the target domain may contain testing data drawn from a significantly different distribution, e.g. different semantic classes than source domain, or simply random noise. We refer to the above testing data as strong out-of-distribution~(strong OOD) data, as opposed to distribution-shifted testing data, e.g. common corruptions, which are referred to as weak OOD data in this work. The ignorance of this realistic setting by existing works, thus, drives us to explore improving the robustness of open-world test-time training~(OWTTT) where testing data is contaminated with strong OOD samples.

\begin{figure*}[!htb]
    \centering
    \includegraphics[width=0.87\linewidth]{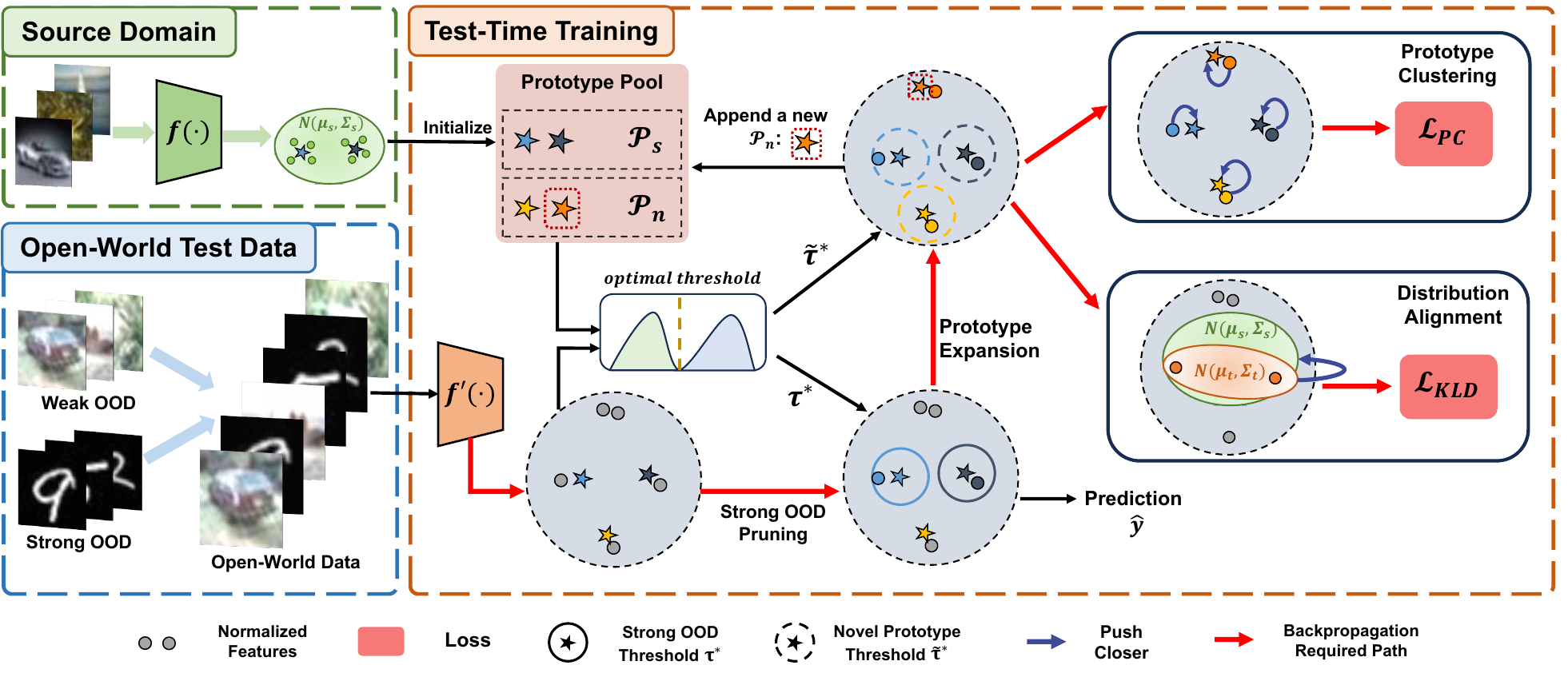}
    \vspace{-0.2cm}
    \caption{An overview of the proposed method for open-world test-time training. Testing data may be contaminated with strong OOD samples. We first prune out potential strong OOD samples by $\tau^*$ for inference. Then, we expand the prototype pool by isolating strong OOD samples and update model weights through self-training and distribution alignment. }
    \label{fig:overview}
    \vspace{-0.3cm}
\end{figure*}

As shown in Fig.~\ref{fig:performance_drop}, we first empirically evaluate existing TTT methods and reveal that TTT methods through self-training~\cite{wang2020tent,liang2020we} and distribution alignment~\cite{YuejiangLiu2021TTTWD,su2022revisiting} 
would all suffer substantially subject to strong OOD samples. These results suggest applying existing TTT techniques fails to achieve safe test-time training in the open-world. We attribute the failure to the following two reasons. First, self-training based TTT~\cite{liang2020we} struggles to deal with strong OOD samples as it has to assign testing samples to known classes. Despite some low confident samples that could be filtered out by applying a threshold as adopted in semi-supervised learning~\cite{sohn2020fixmatch}, it is still not guaranteed to prune out all strong OOD samples. This issue exacerbates when strong OOD samples are not ``strong enough''. For example, when adapting the model pre-trained on CIFAR10 to CIFAR10-C, testing samples drawn from CIFAR100-C are significantly more difficult to be pruned out than random noise samples. Thereby, learning from incorrectly labeled strong OOD samples would weaken the model's discrimination on weak OOD samples. 
Second, distribution alignment based approaches would suffer when strong OOD samples are counted for estimating target domain distribution. Both global distribution alignment~\cite{YuejiangLiu2021TTTWD} and category-wise distribution alignment~\cite{su2022revisiting} could be affected and lead to less accurate feature distribution alignment.

With the potential causes for the failure of existing TTT methods in mind, we propose two techniques to improve the robustness of open-world TTT under a self-training framework.  First, we build the baseline of TTT upon a variant of self-training, i.e. clustering in the target domain with source domain prototypes as cluster centers. To lessen the influence of self-training with incorrectly pseudo-labeled strong OOD samples, we design a hyper-parameter-free method to prune out strong OOD samples. To further separate the features of weak and strong OOD samples, we allow the prototype pool to expand by including isolated strong OOD samples.  Therefore, self-training will allow strong OOD samples to form tight clusters around newly expanded strong OOD prototypes. 
Second, 
as strong OOD samples are likely to be attracted by newly grown prototypes, a more accurate distribution characterizing the target domain data can be estimated. This would benefit distribution alignment between source and target domains. Therefore, we further propose to regularize self-training with global distribution alignment to reduce the risk of confirmation bias~\cite{arazo2020pseudo}. Finally, to synthesize an open-world TTT scenario, we employ CIFAR10-C, CIFAR100-C, ImageNet-C, VisDA-C, ImageNet-R, Tiny-ImageNet, MNIST, and SVHN datasets and create a benchmark by treating one dataset as weak OOD and others as strong OOD. We refer to this benchmark as the open-world test-time training benchmark and hope this would encourage more future works to pay attention to the robustness of test-time training in more realistic scenarios.

We summarize the contributions of this work below.

\begin{itemize}
    
    \item Overlooked by existing studies into test-time training, we argue that open-world test-time training~(OWTTT) could be spoiled by strong OOD testing data. We demonstrated that without special treatment state-of-the-art TTT methods fail to generalize well under the open-world protocol.

    \item  We introduce a baseline method by prototype clustering with distribution alignment regularization. A strong OOD detector and prototype expansion are further developed to improve the robustness of the baseline under OWTTT protocol.

    \item We established a benchmark for evaluating OWTTT protocol covering multiple types of domain shift, including common corruptions and style transfer. Our approach achieves state-of-the-art performance on the proposed benchmark.

\end{itemize}

\section{Related Work}
\label{sec:formatting}

\subsection{Unsupervised Domain Adaptation} 
Unsupervised domain adaptation (UDA) \cite{YaroslavGanin2015UnsupervisedDA} aims to improve models' ability to generalize to target domain data where no labeled data exists. 
UDA is often achieved by learning invariant features across source and target domains~\cite{YaroslavGanin2015UnsupervisedDA}, discovering cluster structures in the target domain\cite{tang2020discriminative, xu2020adversarial, chen2022reusing}, self-supervised training\cite{liu2021cycle}, distance-based alignment\cite{lao2021hypothesis}, etc. Although UDA has made considerable progress in improving the generalizability of the target domain, having access to both the source and target domain during adaptation is not always realistic, e.g. due to data privacy issues. Source-free domain adaptation~(SFDA)~\cite{liang2020we,qiu2021source,ding2022source} has thus emerged which gets rid of the access to source domain data. Nevertheless, SFDA is still not versatile enough to deal with a more realistic scenario where target domain distribution is not known before testing begins.

\subsection{Test-Time training} 
Considering that in some scenarios we would like models that have been deployed to the target domain to automatically adapt to the new environment without accessing source domain data. With these considerations in mind, 
In response to the demand for adaptation to arbitrary unknown target domain with low inference latency, test-time training/adaptation~(TTT/TTA)~\cite{YuSun2019TestTimeTW, wang2020tent} have emerged. 
TTT is often realized by three types of paradigms. Self-supervised learning on the testing data enables adapting to the target domain without considering any semantic information~\cite{YuSun2019TestTimeTW, YuejiangLiu2021TTTWD}. Self-training reinforces model's prediction on unlabeled data and has been demonstrated to be effective for TTT~\cite{wang2020tent,chen2022contrastive,liang2020we,goyaltest2022}. Lastly, distribution alignment provides another viable approach towards TTT by adjusting model weights to produce features following the same distribution as the source domain~\cite{su2022revisiting, YuejiangLiu2021TTTWD}. Despite the efforts into developing more sophisticated TTT methods, certifying the robustness of TTT is still yet to be fully investigated. Recent works studied the robustness of TTT when target domain distribution shifts over time~\cite{wang2022continual,gongnote2022}. A more extensive investigation into TTT under small batch size and imbalanced classes was carried out~\cite{niu2023towards}. Orthogonal to the existing attempts into robustifying TTT, we revealed that TTT is extremely vulnerable to an open-world scenario where testing data consists of strong OOD samples. In this work, we aim to improve the robustness of open-world test-time training by self-training with a dynamic expanded prototype pool.

\subsection{Open-Set Domain Adaptation}
\cite{WalterJScheirer2013TowardOS} introduced the concept of open-set recognition, referring to a setting in that trained models are required to reject testing samples drawn from unknown semantic categories. In the context of domain adaptation, ATI~\cite{PauPanaredaBusto2017OpenSD} proposed open-set domain adaptation~(OSDA) and implemented open-set identification by defining and maximizing the open-set to closed-set distance. OSBP~\cite{KuniakiSaito2018OpenSD} uses backpropagation methods to make the logits of the unknown class samples into a recognizable constant by joint adversarial training of the generator and classifier. DAOD \cite{ZhenFang2019OpenSD} offers a new perspective by proposing unsupervised open-set domain adaptation (UOSDA) where the target domain has unknown classes that are not found in the source domain. \cite{YimingXu2021OpenSD} moved towards a soft rejection method for open-set domain adaptation. In the open-world setting we advocate, we assume testing data could be contaminated by arbitrary strong OOD data, thus it makes little sense to discover new categories for strong OOD data. Instead, we define open-world test-time training as classifying testing samples from source domain categories and rejecting any unknown OOD sample.

\section{Methodology}

In this section, we first overview the formulation of test-time training. Then, we introduce TTT by prototype clustering and how to expand prototypes for open-world test-time training. Distribution alignment is finally combined with prototype clustering to achieve robust open-world test-time training. An overview of the method is presented in Fig.~\ref{fig:overview}.

\subsection{Overview of Test-Time Training}

We first briefly review the practice of test-time training/adaptation for classification tasks. Test-time training aims to adapt the source domain pre-trained model to the target domain which may be subject to a distribution shift from the source domain. We first give an overview of the self-training based TTT paradigm, following the protocol defined in~\cite{su2022revisiting}. In specific, we define the source and target domain datasets as $\set{D}_s=\{x_i,y_i\}_{i=1\cdots N_s}$ with label space $\set{C}_s=\{1\cdots K_s\}$ and $\set{D}_t=\{x_i,y_i\}_{i=1\cdots N_t}$ with label space $\set{C}_t=\{1\cdots K_s, K_s+1, \cdots K_s+K_t\}$. In closed-world TTT, the two label spaces are identical while $\set{C}_s\subseteq\set{C}_t$ is true under open-world TTT. At the testing stage, a minibatch of testing samples at timestamp $t$ is denoted as $\set{B}_t$. We further denote the representation learning network as $z_i=f(x_i;\Theta)\in\mathbbm{R}^D$ and the classifier head as $h(z_i;\omega,\beta)$. Test-time training is achieved by updating the representation network and/or classifier parameters on the target domain dataset $\set{D}_t$. To avoid confusion between TTT definitions, we adopt the sequential test-time training~(sTTT) proposed in \cite{su2022revisiting} for evaluation. Under the sTTT protocol, testing samples are sequentially tested and a model update is carried out after a minibatch of testing samples is observed. The prediction on any testing sample arriving at time-stamp $t$ will not be affected by any testing samples arriving at $t+k$ where $k\geq 1$.

\subsection{TTT by Prototype Clustering}\label{sect:proto_clust}

Inspired by the success of discovering clusters in domain adaptation tasks~\cite{tang2020discriminative, KuniakiSaito2018OpenSD}, we formulate test-time training as discovering cluster structures in the target domain data. The cluster structures are identified in the target domain by identifying representative prototypes as cluster centers and testing samples are encouraged to embed close to one of the prototypes. Inference is then enabled by measuring the testing samples' similarity to the prototypes in the feature space. Formally, we write the prototypes in the source domain as $\set{P}_s=\{p_k\in\mathbbm{R}^{D}\}_{k\in\set{C}_s}$. The prototype clustering objective is defined as minimizing the following negative log-likelihood loss, where $\hat{y}$ indicates the pseudo label $\hat{y}=\arg\max_k h(f(x))$ for $z_i$ and $<\cdot,\cdot>$ measures the cosine similarity.
\vspace{-0.3cm}

\begin{equation}\label{eq:loss_ce}
    \mathcal{L}_{PC}=-\sum\limits_{k\in\set{C}_s}\mathbbm{1}(\hat{y}=k)\log \frac{\exp(\frac{<\vect{p}_k,\vect{z}_i>}{\delta})}{\sum_l\exp(\frac{<\vect{p}_l,\vect{z}_i>}{\delta})}
\end{equation}

 Minimizing the above objective will allow testing samples to embed close to their predicted prototypes and away from other prototypes. In a closed-world test-time training scenario~\cite{su2022revisiting}, prototype clustering has demonstrated strong performance~\cite{liang2020we}. Nevertheless, prototype clustering is severely challenged in the open-world test-time training scenario where strong OOD samples may exist. Should strong OOD samples be forcibly categorized into any source category, self-training on the noisy labeled sample would confuse the network's discriminative capability on weak OOD samples.

\noindent\textbf{Strong OOD Pruning}: We develop a hyper-parameter-free approach to prune out strong OOD samples to avoid the negative impact of adapting model weights. Specifically, we define a strong OOD score $os_i$ for each testing sample as the highest similarity to source domain prototypes as in Eq.~\ref{eq:ood_score}. 

\begin{equation}\label{eq:ood_score}
    os_i = 1-\max\limits_{p_k\in\set{P}_s} <f(x_i),p_k>
\end{equation}

\begin{figure}[t]
    \centering
    \includegraphics[width=0.80\linewidth]{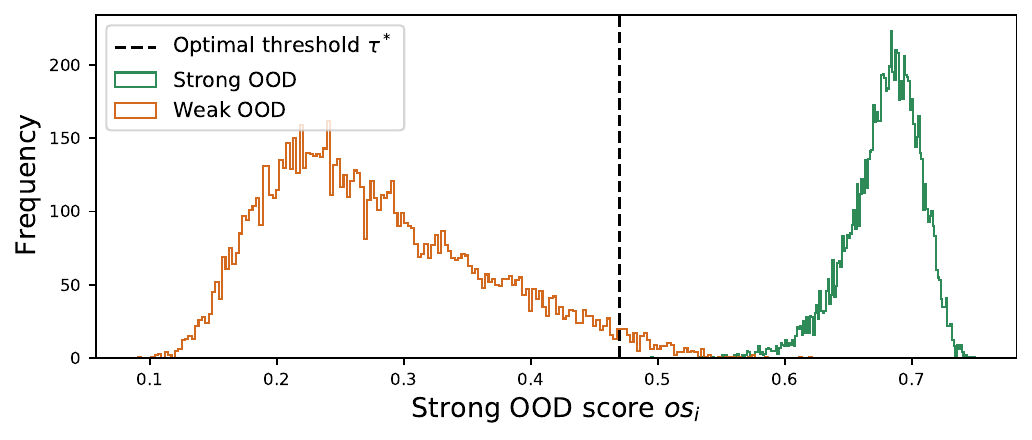}
    \vspace{-0.2cm}
    \caption{A bimodal distribution of strong OOD scores is observed. An optimal threshold separates weak and strong OOD samples.}
    \label{fig:dist_distribution}
    \vspace{-0.5cm}
\end{figure}

We make an observation that the outlier score is subject to a bimodal distribution as shown in Fig.~\ref{fig:dist_distribution}. Therefore, instead of specifying a fixed threshold, we define the optimal threshold as separating the two distribution modalities. In specific, the problem can be formulated as dividing the outlier scores into two clusters and the optimal threshold will minimize the intra-cluster variations in Eq~\ref{eq:otsu_loss}, where $N^+=\sum_i \mathbbm{1}(os_i>\tau)$ and $N^-=\sum_i \mathbbm{1}(os_i\leq\tau)$. Optimizing Eq~\ref{eq:otsu_loss} can be efficiently implemented by exhaustively searching over all possible thresholds from $0$ to $1$ with a step of $0.01$. To maintain a stable estimation of the outlier score distribution, we update the distribution in an exponential moving average manner, with length $N_m$.

\begin{equation}\label{eq:otsu_loss}
\vspace{-0.2cm}
\begin{split}
    \min_\tau \frac{1}{N^+}\sum_i{[os_i-\frac{1}{N^+}\sum_j \mathbbm{1}(os_j>\tau)os_j]^2} + \\
    \frac{1}{N^-}\sum_i{[os_i-\frac{1}{N^-}\sum_j \mathbbm{1}(os_j\leq\tau)os_j]^2}
\end{split}
\vspace{-0.2cm}
\end{equation}

With the optimal threshold $\tau^*$, we could identify strong OOD samples with two benefits.  First, during the 
model weights update, we can exclude detected OOD samples from self-training w.r.t. source domain prototypes. Second, it provides us with a way to differentiate weak OOD samples from strong ones during the inference stage.

\subsection{Open-World TTT by Prototype Expansion}

Identifying strong OOD samples and excluding them from updating model weights does not guarantee a good separation between weak OOD testing samples from strong OOD ones. Inspired by the success of novelty detection~\cite{pimentel2014review}, we propose to dynamically expand the prototype pool to incorporate prototypes representing strong OOD samples. Self-training is then applied with both source domain prototypes and strong OOD prototypes to create wider gaps between weak and strong OOD samples in the feature space. 
Specifically, we denote 
additionally a strong OOD prototype set as $\set{P}_n=\{p_k\in\mathbbm{R}^D\}_{k\in\set{C}_t\setminus\set{C}_s}$. When no prior information on the target domain is available, we initialize the novel prototypes as an empty set. As TTT goes on, $\set{P}_n$ is expected to expand to accommodate the unknown distribution in the target domain.

\noindent\textbf{Prototype Expansion}: Expanding a strong OOD prototype pool requires evaluating testing samples against both source domain and strong OOD prototypes. A similar problem was investigated for the purpose of dynamically estimating the number of clusters from data. A deterministic hard clustering algorithm, DP-means~\cite{kulis2011revisiting}, was developed by measuring the distance of data points to known cluster centers, and a new cluster will be initialized when the distance is above a threshold. DP-means is shown to be equivalent to optimizing a K-means objective with an additional penalty on the number of clusters. DP-means was later adapted to few-shot learning tasks~\cite{allen2019infinite}. DP-means provides a viable solution to dynamic prototype expansion with an additional threshold hyper-parameter $\tau$, which can be estimated from the concentration parameter $\alpha$ for the Chinese restaurant process~\cite{aldous1985exchangeability}. Despite the connection to theoretical stochastic processing, estimating $\tau$ requires knowing the standard deviation for the base distribution which is still not trivial. To ease the difficulty of estimating additional hyper-parameter, we first define a testing sample with an extended strong OOD score $\tilde{os}_i$ as the closest distance to existing source domain prototypes and strong OOD prototypes as in Eq.~\ref{eq:ext_ood_score}.
A dynamic threshold is estimated following Eq.~\ref{eq:otsu_loss}. Hence, testing samples above this threshold will establish a new prototype. To avoid adding close-by testing samples, we repeat the additive process incrementally. Since model weights are subject to constant updates, OOD prototypes may experience a representation shift. To avoid expanding OOD prototypes indefinitely, we treat the $\set{P}_n$ as a queue with fixed length $N_q$, and old OOD prototypes will be discarded when new OOD prototypes are appended.

\begin{equation}\label{eq:ext_ood_score}
% \begin{split}
    \tilde{os}_i=1-\max\limits_{p_k\in\set{P}_{s}\bigcup\set{P}_n}<f(x_i),p_k>
% \end{split}
\end{equation}

\noindent\textbf{Prototype Clustering with Strong OOD Prototypes}: With additional strong OOD prototypes being identified, we define the prototype clustering loss for testing samples with two considerations. First, testing samples classified as known classes should embed closer to the prototype and keep away from other prototypes, which defines a K-way classification task. Second, testing samples classified as strong OOD prototypes should embed further away from any source domain prototypes, which defines a K+1-way classification task. With these objectives in mind, we define the prototype clustering loss as in Eq.~\ref{eq:protoype_clust}, where the pseudo labels are predicted as $\hat{y}_i=\arg\max_{p_k\in\set{P}_s\bigcup\set{P}_n}<p_k,z_i>$.

\begin{equation}\label{eq:protoype_clust}
\begin{split}
    \mathcal{L}_{PC} = -\sum_{k\in\set{C}_s} \mathbbm{1}(\hat{y}_i=k)\log\frac{\exp(\frac{<p_k,z_i>}{\delta})}{\sum\limits_{l\in\set{C}_s}\exp(\frac{<p_l,z_i>}{\delta})}\\
    - \sum_{k\in\set{C}_t} \mathbbm{1}(\hat{y}_i=k)\log\frac{\exp(\frac{<p_k,z_i>}{\delta})}{\sum\limits_{l\in\set{C}_s+1}\exp(\frac{<p_l,z_i>}{\delta})}
\end{split}
\end{equation}

\subsection{Distribution Alignment Regularization}

Self-training~(ST) is known to be prone to incorrect pseudo labels, a.k.a. confirmation bias~\cite{arazo2020pseudo}. The situation exacerbates when the target domain consists of OOD samples. To reduce the risk of ST failure, we further incorporate distribution alignment~\cite{YuejiangLiu2021TTTWD} as regularization to self-training. Specifically, we assume a Gaussian distribution $\mathcal{N}(\mu_s,\Sigma_s)$ for the source domain feature. In the target domain, we incrementally estimate the feature distribution $\mathcal{N}(\mu_t,\Sigma_t)$ with a momentum $\beta$~\cite{su2022revisiting}. Only testing samples passing the strong OOD pruning are included for estimating the distribution. The KL-Divergence loss $\mathcal{L}_{KLD}$ between source and target domain distributions is finally utilized to regularize prototype clustering on the target domain.

\begin{equation}\label{eq:distribution_align}
    \mathcal{L}_{KLD}=D_{KL}(\mathcal{N}(\mu_s,\Sigma_s)||\mathcal{N}(\mu_t,\Sigma_t))
\end{equation}

\subsection{Open-World TTT Algorithm}

In this section, we provide an overview of the open-world TTT algorithm in Algo.~\ref{alg:main}. We divide the TTT procedure into two steps, the inference stage and the adaption stage. 

\noindent\textbf{Inference Stage}: i) We evaluate the testing samples' strong OOD score $os_i$ by Eq.~\ref{eq:ood_score} and categorize the samples into weak or strong OOD. ii) We further categorize weak OOD samples into one of the $\set{C}_s$ classes, strong OOD samples are simply counted as one separate strong OOD class.

\noindent\textbf{Adaptation Stage}: i) We calculate the extended strong OOD score and threshold and expand the prototype pool by the isolated strong OOD sample. ii) We calculate test-time training loss $\mathcal{L}_{PC}$ and $\mathcal{L}_{KLD}$ and update model by gradient descent.
For test-time training on the full testing data, we repeat the two stages until the testing data is exhausted.

\vspace{-0cm}
\begin{algorithm}

\setstretch{1.2}
\caption{Open-World TTT Algorithm }\label{alg:main}

\SetKwInOut{Input}{input}
\SetKwInOut{Return}{return}
\Input{Testing sample batch $\set{B}^t=\{x_i\}_{i=1\cdots N_B}$}

\textcolor{gray}{\# Inference Stage:}\\
\For{$x_i \leftarrow 1$ \KwTo $N_B$}
{
 Calculate $os_i$ \& $\tau^*$ by Eq.~\ref{eq:ood_score} \& \ref{eq:otsu_loss}\;

\textcolor{gray}{\# Label Prediction}:\\
\eIf{$os_i < \tau^*$}
{$\hat{y}_i=\arg\max_{p_k\in \set{P}_s} <f(x_i;\Theta),p_k>$}
{$\hat{y}_i=|\set{C}_s|+1$}
}

\textcolor{gray}{\# Adaptation Stage:}\\
\For{$x_i \leftarrow 1$ \KwTo $N_B$}
{
 Calculate $\tilde{os}_i$ \& $\tilde{\tau}^*$ by Eq.~\ref{eq:ext_ood_score} \& \ref{eq:otsu_loss}  \;

\textcolor{gray}{\# Prototype Expansion:}\\
\If{$\tilde{os}_i > \tilde{\tau}^*$}
{ $\set{P}_n=\set{P}_n\bigcup f(x_i;\Theta)$}
}

\For{$x_i \leftarrow 1$ \KwTo $N_B$}
{
 Calculate losses $\set{L}_{PC}$ \& $\set{L}_{KLD}$ by Eq.~\ref{eq:protoype_clust} \& \ref{eq:distribution_align}\;

 Update $\Theta=\Theta - \alpha (\nabla\set{L}_{PC}+\lambda\nabla\set{L}_{KLD})$\;
}
\end{algorithm}
\vspace{-0.3cm}

\begin{table*}[t]
\caption{Open-world test time training results on CIFAR10-C. All numbers are in $\%$.}\label{tab:CIFAR10-C}
\vspace{-0.2cm}
\scriptsize
\centering{
\begin{tabular}{l  ccc  ccc  ccc  ccc  ccc }
\toprule
\multirow{2}*{Method}& \multicolumn{3}{c}{Noise}& \multicolumn{3}{c}{MNIST}& \multicolumn{3}{c}{SVHN}& \multicolumn{3}{c}{Tiny-ImageNet}& \multicolumn{3}{c}{CIFAR100-C}\\
\cmidrule{2-16}
&$Acc_S$&$Acc_N$&$Acc_H$&$Acc_S$&$Acc_N$&$Acc_H$&$Acc_S$&$Acc_N$&$Acc_H$&$Acc_S$&$Acc_N$&$Acc_H$&$Acc_S$&$Acc_N$&$Acc_H$\\

\cmidrule(lr){2-4}\cmidrule(lr){5-7}\cmidrule(lr){8-10}\cmidrule(lr){11-13}\cmidrule(lr){14-16}

TEST&68.59&\textbf{99.97}&81.36&60.48&88.81&71.96&60.94&86.44&71.48&57.41&79.63&66.72&52.74&74.24&61.67\\
BN&\underline{76.63}&95.69&\underline{85.11}&76.15&\underline{95.75}&\underline{84.83}&\underline{79.18}&\textbf{94.71}&\underline{86.25}&67.66&\underline{82.67}&74.42&68.44&\underline{81.38}&\underline{74.35}\\
TTT++&41.09&57.31&47.86&59.52&77.52&67.34&68.77&85.80&76.34&66.70&79.28&72.44&65.69&77.47&71.10\\
TENT&32.24&33.30&32.77&55.64&68.27&61.31&66.70&82.50&73.77&66.54&79.32&72.37&64.80&76.40&70.12\\
SHOT&63.54&71.37&67.23&56.92&53.26&55.03&70.01&72.58&71.27&67.78&82.25&\underline{74.32}&67.73&72.87&70.21\\
TTAC&64.46&77.42&70.35&\underline{77.60}&84.53&80.92&77.30&81.10&79.16&\underline{71.64}&77.14&74.29&\underline{71.94}&75.44&73.65\\
OURS&\textbf{85.46}&\underline{98.60}&\textbf{91.56}&\textbf{83.89}&\textbf{97.83}&\textbf{90.32}&\textbf{84.99}&\underline{87.94}&\textbf{86.44}&\textbf{71.77}&\textbf{84.71}&\textbf{77.70}&\textbf{74.08}&\textbf{84.64}&\textbf{79.01}\\
\bottomrule
\end{tabular}}
\vspace{-0.3cm}
\end{table*}

\begin{table*}[t]
\caption{Test time training results on CIFAR100-C.}\label{tab:CIFAR100-C}
\vspace{-0.2cm}
\scriptsize\centering{

\begin{tabular}{l  ccc  ccc  ccc  ccc  ccc }
\toprule
\multirow{2}*{Method}& \multicolumn{3}{c}{Noise}& \multicolumn{3}{c}{MNIST}& \multicolumn{3}{c}{SVHN}& \multicolumn{3}{c}{Tiny-ImageNet}& \multicolumn{3}{c}{CIFAR10-C}\\
\cmidrule{2-16}
&$Acc_S$&$Acc_N$&$Acc_H$&$Acc_S$&$Acc_N$&$Acc_H$&$Acc_S$&$Acc_N$&$Acc_H$&$Acc_S$&$Acc_N$&$Acc_H$&$Acc_S$&$Acc_N$&$Acc_H$\\
\cmidrule(lr){2-4}\cmidrule(lr){5-7}\cmidrule(lr){8-10}\cmidrule(lr){11-13}\cmidrule(lr){14-16}

TEST&36.75&\textbf{99.87}&53.73&25.99&49.59&34.11&30.01&81.62&43.89&25.41&70.06&37.3&25.55&73.28&37.89\\
BN&50.21&\underline{98.72}&66.56&36.21&84.69&50.73&45.69&\underline{90.45}&60.71&34.88&\textbf{82.18}&48.97&37.00&\underline{83.54}&51.28\\
TTT++&23.47&70.26&35.19&28.31&\textbf{86.74}&42.68&37.56&90.45&53.08&34.67&81.25&48.60&33.78&81.12&47.70\\
TENT&22.57&66.60&33.72&27.85&80.92&41.43&37.08&89.90&52.51&35.51&77.34&48.68&35.20&80.26&48.94\\
SHOT&\underline{51.52}&98.21&\underline{67.58}&35.35&81.71&49.35&45.87&89.72&60.70&\underline{35.72}&81.11&\underline{49.59}&38.00&82.13&51.96\\
TTAC&51.11&98.66&67.34&\underline{37.78}&\underline{86.66}&\underline{52.62}&\underline{47.29}&\textbf{91.42}&\underline{62.33}&32.04&80.46&45.83&\underline{38.83}&\textbf{83.68}&\underline{53.05}\\
OURS&\textbf{56.76}&97.25&\textbf{71.68}&\textbf{40.77}&82.91&\textbf{54.66}&\textbf{54.32}&81.98&\textbf{65.34}&\textbf{38.90}&\underline{81.92}&\textbf{52.75}&\textbf{38.97}&83.20&\textbf{53.08}\\
\bottomrule
\end{tabular}}
\vspace{-0.3cm}
\end{table*}

\begin{table}[th!]
\caption{Test time training results on ImageNet-C.}\label{tab:ImageNet-C}
\vspace{-0.2cm}
\centering{
\setlength\tabcolsep{5pt} 
\resizebox{0.99\linewidth}{!}{
\begin{tabular}{l  ccc  ccc  ccc }
\toprule
\multirow{2}*{Method}& \multicolumn{3}{c}{noise}& \multicolumn{3}{c}{MNIST}& \multicolumn{3}{c}{SVHN}\\
\cmidrule{2-10}
&$Acc_S$&$Acc_N$&$Acc_H$&$Acc_S$&$Acc_N$&$Acc_H$&$Acc_S$&$Acc_N$&$Acc_H$\\
\cmidrule(lr){2-4}\cmidrule(lr){5-7}\cmidrule(lr){8-10}

TEST&18.51&\textbf{100.00}&31.24&18.66&\textbf{98.27}&31.36&18.94&\underline{87.75}&31.15\\
BN&36.34&99.97&53.31&\underline{30.77}&74.53&\underline{43.55}&33.26&84.54&\underline{47.74}\\
TENT&22.54&10.47&14.29&27.53&10.01&14.68&\underline{41.16}&45.51&43.22\\

SHOT&\textbf{46.79}&\textbf{100.00}&\textbf{63.75}&27.47&55.25&36.70&34.00&75.94&46.97\\

TTAC&\underline{42.60}&94.52&\underline{58.73}&30.43&72.11&42.80&31.59&74.07&44.29\\
OURS&41.40&\textbf{100.00}&58.56&\textbf{38.86}&\underline{93.35}&\textbf{54.87}&\textbf{38.60}&\textbf{98.06}&\textbf{55.40}\\
\bottomrule
\end{tabular}}
}
\vspace{-0.3cm}
\end{table}

\section{Experiments}

We validate the effectiveness of open-world TTT and extensively benchmarked multiple state-of-the-art TTT methods. In the following of this section, we first elaborate on the datasets, evaluation metrics, and competing methods. Then, we present open-world TTT results on five datasets and analyze the effectiveness of individual components.

\subsection{Datasets}

We introduce several datasets, including corruption datasets, style transfer datasets, and some other common datasets. For the corruption datasets, We selected \textbf{CIFAR10-C/CIFAR100-C} \cite{hendrycks2019benchmarking} as a small corruption dataset, each containing 10,000 corrupt images with 10/100 categories, and \textbf{ImageNet-C} \cite{hendrycks2019benchmarking} as a large-scale corruption dataset, which contains 50,000 corruption images within 1000 categories. We also introduced some style transfer datasets. \textbf{VisDA-C} \cite{peng2017visda} is a synthetic to real large-scale dataset containing 152,397 synthetic training images and 55,388 real testing images, belonging to 12 categories. \textbf{ImageNet-R} \cite{hendrycks2021many} is a large-scale realistic style transfer dataset that has renditions of 200 ImageNet classes resulting in 30,000 images. \textbf{Tiny-ImageNet} \cite{le2015tiny} consists of 200 categories with each category containing 500 training images and 50 validation images. We also introduce some digits datasets.  \textbf{MNIST} \cite{lecun1998gradient} is a handwritten digit dataset, which contains 60,000 training images and 10,000 testing images. \textbf{SVHN} \cite{netzer2011reading} is a digital dataset in a real street context, including 50,000 training images and 10,000 testing images.

\subsection{Evaluation Metric}
To evaluate open-world test-time training, we adopt an evaluation metric similar to open-set classification tasks~\cite{geng2020recent}. The major concern for TTT evaluation is the accuracy of source domain categories and whether strong OOD samples can be rejected. Therefore, we define two accuracy metrics, the first measures the accuracy of source domain classes, denoted $Acc_S$, which is equivalent to existing TTT metrics. We further define the accuracy on strong OOD categories, denoted as $Acc_N$, as a binary classification task, i.e. true positive is defined as successfully rejecting a strong OOD sample as any of the source domain classes. As we pursue a good trade-off between source accuracy and OOD accuracy, we further report a harmonic mean $Acc_H$ between $Acc_S$ and $Acc_N$ to measure the balanced prediction. More detailed definitions are given in Eq.~\ref{eq:metrics}, where $\hat{y}_i$ refers to the predicted label and $\mathbbm{1}(y_i\in\set{C}_s)$ is true if $y_i$ is in the set $\set{C}_s$.

\begin{equation}\label{eq:metrics}
\vspace{-0.3cm}
\begin{split}
    &Acc_S=\frac{\sum_{x_i,y_i\in\set{D}_t}\mathbbm{1}(y_i=\hat{y}_i)\cdot\mathbbm{1}(y_i\in\set{C}_s)}{\sum_{x_i,y_i\in\set{D}_t}\mathbbm{1}(y_i\in\set{C}_s)}\\
    &Acc_N=\frac{\sum\limits_{x_i,y_i\in\set{D}_t}\mathbbm{1}(\hat{y}_i\in\set{C}_t\setminus\set{C}_s)\cdot\mathbbm{1}(y_i\in\set{C}_t\setminus\set{C}_s)}{\sum_{x_i,y_i\in\set{D}_t}\mathbbm{1}(y_i\in\set{C}_t\setminus\set{C}_s)}\\
    &Acc_H=2\cdot\frac{Acc_S\cdot Acc_N}{Acc_S+Acc_N}
\end{split}
\end{equation}

\subsection{Open-World Test-Time Training Protocol}
We propose the Open-World Test Time Training (OWTTT) protocol to evaluate different TTT methods' performance in the open-world case. In order to set up a fair comparison with existing methods, we take all the classes in the TTT benchmark dataset as seen classes and add additional classes from additional datasets as unseen classes. By doing so we do not need to modify the source domain training process. In the later experiments, we set the number of known class samples and the number of unknown class samples to be the same. Then we follow the "One Pass" protocol in \cite{su2022revisiting}, which has two main restrictions. Firstly, the training objective cannot be changed during the source domain training procedure (e.g. add additional self-training branches). Secondly, testing data in the target domain is sequentially streamed and predicted.

\subsection{Training Details}

We followed the sequential test-time training protocol specified in~\cite{su2022revisiting} and choose ResNet-50 \cite{KaimingHe2022DeepRL} as the backbone network for all experiments. For optimization, we choose SGD with momentum to optimize the backbone network. We set learning rate $\alpha$=\{1e-3, 1e-4, 2.5e-5, 2.5e-5, 2.5e-5\}, batch size $N_B=\{256, 256, 128, 128, 128\}$, $\lambda=\{1, 1, 0.4, 0.4, 0.04\}$, respectively for experiments on Cifar10-C, Cifar100-C, ImageNet-C, ImageNet-R and VisDA-C, respectively. To further reduce the effect of incorrect pseudo labeling, we only use 50$\%$ of samples with $od_i$ far from $\tau^*$ to perform prototype clustering for each batch. For all experiments, we use temperature scaling $\delta$ = 0.1,  the length of strong OOD prototypes queue $N_q$ = 100, and the length of moving average $N_m$ = 512.

\subsection{Competing Methods}

As no existing TTT methods have considered the challenges in open-world TTT, we benchmark the following test-time training methods. \textbf{TEST} directly evaluates the source domain model on testing data. \textbf{BN} \cite{ioffe2015batch} update batch norm statistics on the testing data for test-time adaptation. \textbf{TTT++} \cite{YuejiangLiu2021TTTWD} aligns source and target domain distribution by minimizing the F-norm between the mean covariance. \textbf{TENT} \cite{wang2020tent} updates model batch norm affine parameters by minimizing the entropy loss on testing data. \textbf{SHOT} \cite{liang2020we} implements test-time training by entropy minimization and self-training. SHOT assumes the target domain is class balanced and introduced an entropy loss to encourage uniform distribution of the prediction results. \textbf{TTAC} \cite{su2022revisiting} employs distribution alignment at both global and category levels to facilitate test-time training. Finally, we present our method ~(\textbf{Ours}) which combines self-training with prototype expansion to accommodate the strong OOD samples.
For all competing methods, by default, we equip them with the same strong OOD detector introduced in Sect.~\ref{sect:proto_clust}.

\subsection{Evaluation of Open-World TTT}
\noindent\textbf{Open-World TTT on Noise Corrupted Target Domain}:

We first evaluate open-world test-time training under noise corrupted target domain. We treat CIFAR10, CIFAR100 \cite{krizhevsky2009learning} and ImageNet \cite{deng2009imagenet} as the source domain and test-time adapt to CIFAR10-C, CIFAR100-C, and ImageNet-C as the target domain respectively. For experiments on CIFAR10/100, we introduce random noise, MNIST, SVHN, Tiny-ImageNet with non-overlap categories, and CIFAR100 as strong OOD testing samples. For ImageNet-C, we introduce random noise, MNIST, and SVHN as strong OOD samples. The results are presented in Tab.~\ref{tab:CIFAR10-C}, Tab.~\ref{tab:CIFAR100-C} and Tab.~\ref{tab:ImageNet-C} respectively. We make the following observations from the results. First, direct testing does not necessarily gives the worst performance when strong OOD samples are drawn from random noise distribution. This suggests the impact of strong OOD samples is more severe than some corruptions on target domain testing data. Second, BN and TTAC perform relatively well compared with TENT and SHOT across different types of strong OOD testing samples. This result indicates that test-time adaptation through distribution alignment is in general more robust than self-training based methods under the interference of strong OOD testing samples. Finally, our proposed method consistently outperforms all competing methods under most experiment settings, suggesting the effectiveness of the proposed method. In addition to vertically comparing different competing methods, we further notice that the performance variation among different methods is significantly higher on using random noise and MNIST as strong OOD samples than using Tiny ImageNet and CIFAR100 as strong OOD samples. We owe the difference to the fact that there is a less significant distribution shift between CIFAR10 and CIFAR100 than between CIFAR10 and random noise/MNIST.

\noindent\textbf{Open-World TTT on Style Transfer Target Domain}: We further evaluate VisDA-C and ImageNet-R to demonstrate the effectiveness of the style transfer target domain. For both ImageNet-R and VisDA-C the strong OOD samples are selected in a similar way to ImageNet-C and the results are presented in Tab.~\ref{tab:ImageNet-R} and Tab.~\ref{tab:VisDA-C}. We draw similar conclusions from the results as with common corruptions. Our proposed method consistently outperforms competing methods.

\begin{table}[t!]
\caption{Test time training results on ImageNet-R.}\label{tab:ImageNet-R}
\centering{
\resizebox{0.99\linewidth}{!}{
\begin{tabular}{l  ccc  ccc  ccc }
\toprule
\multirow{2}*{Method}& \multicolumn{3}{c}{noise}& \multicolumn{3}{c}{MNIST}& \multicolumn{3}{c}{SVHN}\\
\cmidrule{2-10}
&$Acc_S$&$Acc_N$&$Acc_H$&$Acc_S$&$Acc_N$&$Acc_H$&$Acc_S$&$Acc_N$&$Acc_H$\\
\cmidrule(lr){2-4}\cmidrule(lr){5-7}\cmidrule(lr){8-10}
TEST&35.83&\textbf{100.00}&52.76&35.50&99.96&52.39&35.75&\underline{98.32}&\underline{52.43}\\
BN&40.05&\textbf{100.00}&57.19&39.86&\underline{99.98}&57.00&\underline{38.16}&69.41&49.25\\
TENT&32.35&94.96&48.26&18.77&31.70&23.58&29.17&65.12&40.29\\
SHOT&39.20&\textbf{100.00}&56.32&37.30&96.30&53.77&32.37&57.53&41.43\\

TTAC&\underline{41.24}&\textbf{100.00}&\underline{58.40}&\underline{40.55}&99.41&\underline{57.60}&36.68&68.48&47.77\\
Paper&\textbf{41.66}&\textbf{100.00}&\textbf{58.82}&\textbf{41.40}&\textbf{100.00}&\textbf{58.56}&\textbf{41.47}&\textbf{99.90}&\textbf{58.61}\\
\bottomrule
\end{tabular}}
}
\end{table}

\begin{table}[t!]
\caption{Test time training results on VisDA-C.}\label{tab:VisDA-C}
\centering{
\resizebox{0.99\linewidth}{!}{
\begin{tabular}{l  ccc  ccc  ccc }
\toprule
\multirow{2}*{Method}& \multicolumn{3}{c}{noise}& \multicolumn{3}{c}{MNIST}& \multicolumn{3}{c}{SVHN}\\
\cmidrule{2-10}
&$Acc_S$&$Acc_N$&$Acc_H$&$Acc_S$&$Acc_N$&$Acc_H$&$Acc_S$&$Acc_N$&$Acc_H$\\
\cmidrule(lr){2-4}\cmidrule(lr){5-7}\cmidrule(lr){8-10}
TEST&48.22&\textbf{100.00}&65.07&48.59&\textbf{98.80}&\underline{65.14}&49.38&\textbf{98.48}&64.95\\
BN&52.58&\textbf{100.00}&68.92&50.11&82.34&62.30&\underline{58.49}&\underline{86.19}&\textbf{69.69}\\
TENT&58.80&98.99&73.78&46.27&68.19&55.13&51.00&81.48&62.73\\
SHOT&\underline{62.09}&\textbf{100.00}&\underline{76.61}&48.19&36.00&41.21&56.47&62.84&59.48\\

TTAC&59.21&\textbf{100.00}&74.38&\underline{52.04}&67.38&58.72&57.94&66.27&61.83\\
OURS&\textbf{63.93}&\textbf{100.00}&\textbf{78.00}&\textbf{64.18}&\underline{90.96}&\textbf{75.26}&\textbf{64.76}&74.80&\underline{69.42}\\
\bottomrule
\end{tabular}}
}
\vspace{-0.3cm}
\end{table}

\subsection{Ablation Study}

We investigate the effectiveness of the components in the proposed method. In particular, we evaluate the effectiveness of prototype clustering~(Pro. Cls.), OOD detection~(OOD~Det.), prototype expansion~(Pro.~Exp.), and finally distribution alignment regularization~(Dist.~Al.). We report TTT performance as $Acc_H$ on CIFAR10-C as weak OOD and MNIST, SVHN, and CIFAR100-C as strong OOD in Tab.~\ref{tab:ablation}. When strong OOD detection~(O.D.) is not present, all testing samples are classified as source domain classes, thus resulting in $Acc_N=0\%$. When strong OOD detector~(O.D.) is enabled for inference, we observe a drop of source domain accuracy $Acc_S$ with significantly improved strong OOD accuracy $Acc_N$. We further incorporate prototype clustering~(P.C.) on the target domain by updating source domain prototypes $\set{P}_s$ only. As no strong OOD prototypes are present, direct self-training with source domain prototypes does not necessarily improve $Acc_S$ and $Acc_N$ simultaneously. Therefore, as we include prototype expansion~(P.E) to dynamically expand strong OOD prototypes, consistent improvements upon O.D. without adaptation are observed on the three datasets. Finally, as we combine distribution alignment~(D.A) to regularize self-training, we achieve the best performance on all categories of accuracies. The combined approach also significantly outperforms distribution alignment with strong OOD detection alone.

\begin{table}[t]

  \centering
  \caption{Ablation study on CIFAR10-C as weak OOD samples. We investigate the effectiveness of OOD Detection~(O.D.), Prototype Clustering~(P.C.), Prototype Expansion~(P.E.) and Distribution Alignment~(D.A.).}
  \setlength\tabcolsep{2pt} 
\resizebox{1.01\linewidth}{!}{
    \begin{tabular}{ccccccccccccc}
    \toprule
    \multirow{2}[2]{*}{O.D.} & \multirow{2}[2]{*}{P.C.} & \multirow{2}[2]{*}{P.E.} & \multirow{2}[2]{*}{D.A.} & \multicolumn{3}{c}{Noise} & \multicolumn{3}{c}{SVHN} & \multicolumn{3}{c}{CIFAR100-C} \\
\cmidrule(lr){5-13}      &       &       &       & $Acc_S$ & $Acc_N$ & $Acc_H$ & $Acc_S$ & $Acc_N$ & $Acc_H$ & $Acc_S$ & $Acc_N$ & $Acc_H$ \\
\cmidrule(lr){1-4}\cmidrule(lr){5-7}\cmidrule(lr){8-10}\cmidrule(lr){11-13} 
    -     & -     & -     & -     & 70.6  & 0.0   & 0.0   & 70.6  & 0.0   & 0.0   & 70.6  & 0.0   & 0.0 \\
    \checkmark & -     & -     & -     & 68.6  & 100.0 & 81.4  & 60.9  & 86.4  & 71.5  & 52.7  & 74.2  & 61.7 \\
    \checkmark & $\set{P}_s$ & -     & -     & 65.2  & 91.5  & 76.1  & 60.9  & 90.0  & 72.7  & 56.3&69.0&62.0 \\
    \checkmark & $\set{P}_s$+$\set{P}_n$ & \checkmark & -     & 68.7  & 99.8  & 81.4  & 65.3  & 95.0  & 77.4  & 52.6  & 78.9  & 63.2 \\
    \checkmark & -     & -     & \checkmark & 72.9  & 88.8  & 80.1  & 78.1  & 88.0  & 82.8  & 70.5  & 78.7  & 74.4 \\
    \checkmark & $\set{P}_s$+$\set{P}_n$
    & \checkmark & \checkmark & 85.5  & 98.6  & 91.6  & 85.0  & 87.9  & 86.4  & 74.1  & 84.6  & 79.0 \\
    \bottomrule
    \end{tabular}%
    }
    \vspace{-0.2cm}
  \label{tab:ablation}
\end{table}

\begin{table}[t]
\caption{Evaluating the robustness of our method on CIFAR10-C under different ratios of strong to weak OOD samples.}\label{tab:ratio}
\scriptsize\centering{
\begin{tabular}{l  c  c  c  c  c }
\toprule
Ratio& noise& MNIST& SVHN& Tiny-ImageNet& CIFAR100-C\\
\midrule

0.2&90.83&89.59&86.91&77.64&69.15\\
0.4&91.59&90.55&88.51&77.67&76.99\\
0.6&91.80&90.92&88.28&77.68&78.77\\
0.8&91.70&90.19&86.65&77.42&79.58\\
1.0&91.34&90.34&86.91&77.70&79.04\\
\bottomrule
\end{tabular}}
\vspace{-0.3cm}
\end{table}

\subsection{Additional Analysis}

In this section, we provide additional analysis of open-world TTT from several aspects and explore alternative designs of the model. The performance of open-world TTT is also investigated under different testing data compositions.

\begin{figure}[!htb]
    \centering
    \includegraphics[width=0.85\linewidth]{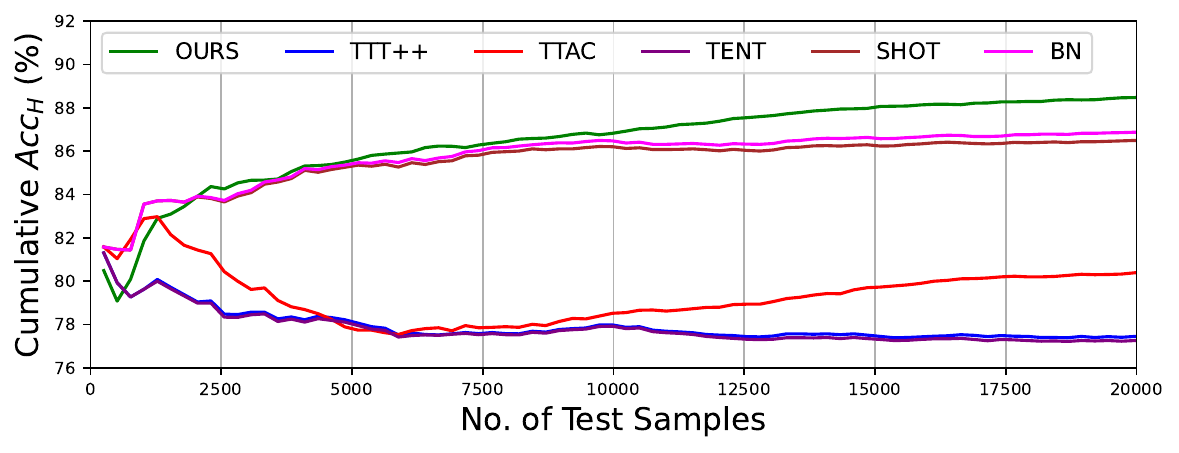}
    \vspace{-0.3cm}
    \caption{Comparison of test-time cumulative $Acc_{H}$. 
    \vspace{-0.3cm}
    }
    \label{fig:cumulative_acc}
\end{figure}

\noindent\textbf{Cumulative performance in the TTT process}: 
We first present the cumulative testing performance under open-world TTT protocol in Fig.~\ref{fig:cumulative_acc}. We report $Acc_H$ for multiple TTT methods. Our method consistently outperforms the competing methods during the whole TTT procedure, and the performance gets progressively better as the number of test samples increases.

\noindent\textbf{Performance under different open world data ratios}:
In real-world applications, the ratio of strong OOD to weak OOD in open-world test data is variable. We examine the impact of different ratios between weak and strong OOD samples on open-world TTT performance. Specifically, we control the ratio between strong to weak OOD samples from 0.2 to 1.0. $ACC_H$ on CIFAR10-C are presented in Tab.~\ref{tab:ratio}. Experiments show that our approach is not sensitive to the data ratio and can be applied to a variety of data ratio scenarios.

\begin{figure}[!htb]
    \vspace{-0.3cm}
    \centering
    \includegraphics[width=0.85\linewidth]{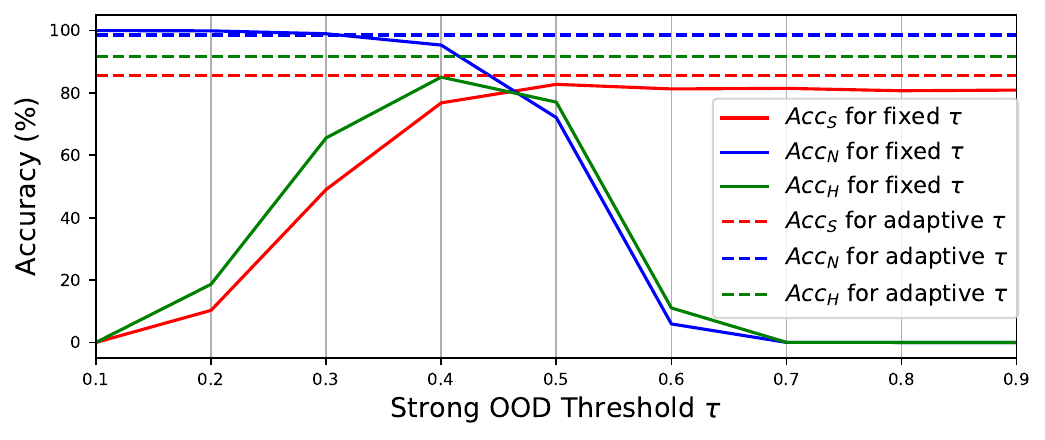}
    \caption{Comparison between the optimal strong OOD threshold $\tau^*$ (dotted lines) and fixed threshold $\tau$ (solid lines) range from 0.1 to 0.9.}
    \label{fig:comparethreshold}
        \vspace{-0.4cm}
\end{figure}

\begin{figure}[!t]
    \vspace{-0.6cm}
    \centering
    \subfloat[TEST]{\includegraphics[width=0.49\linewidth]{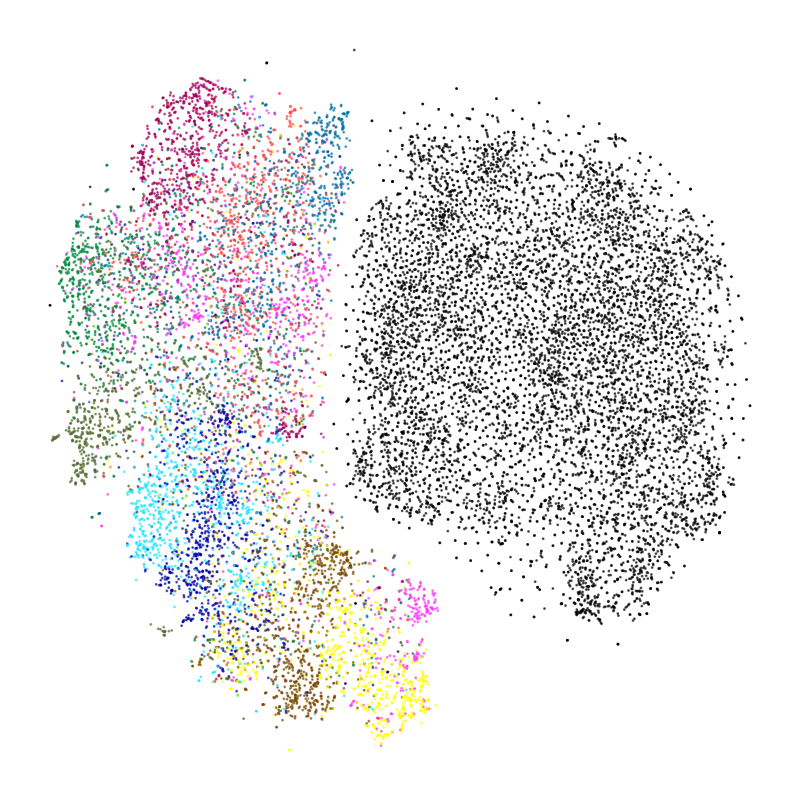}}
    \subfloat[TTAC w/ OOD Det.]{\includegraphics[width=0.49\linewidth]{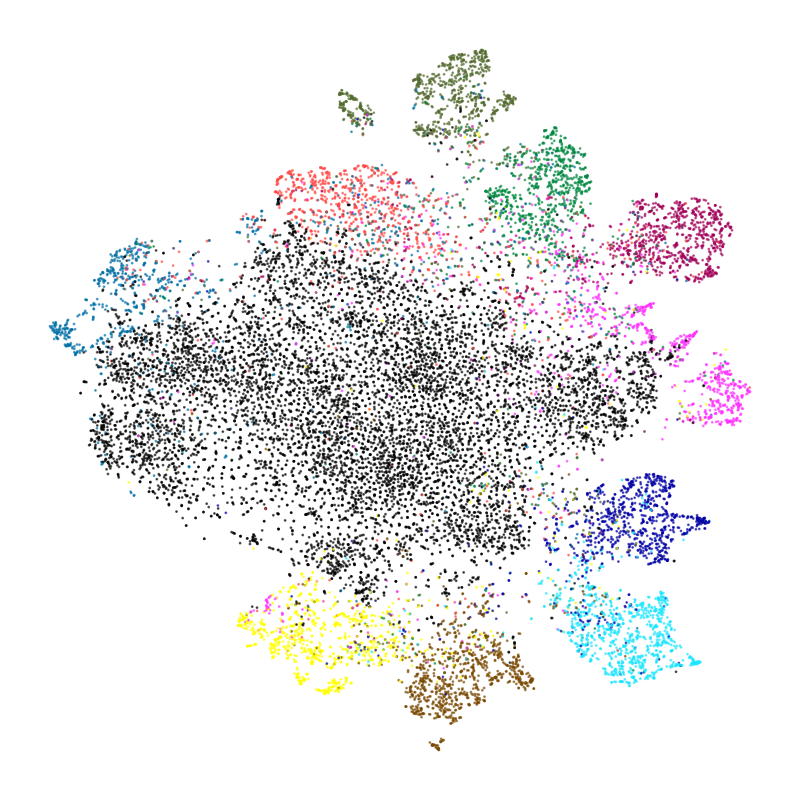}}\\
    \subfloat[TTAC w/o OOD Det.]{\includegraphics[width=0.49\linewidth]{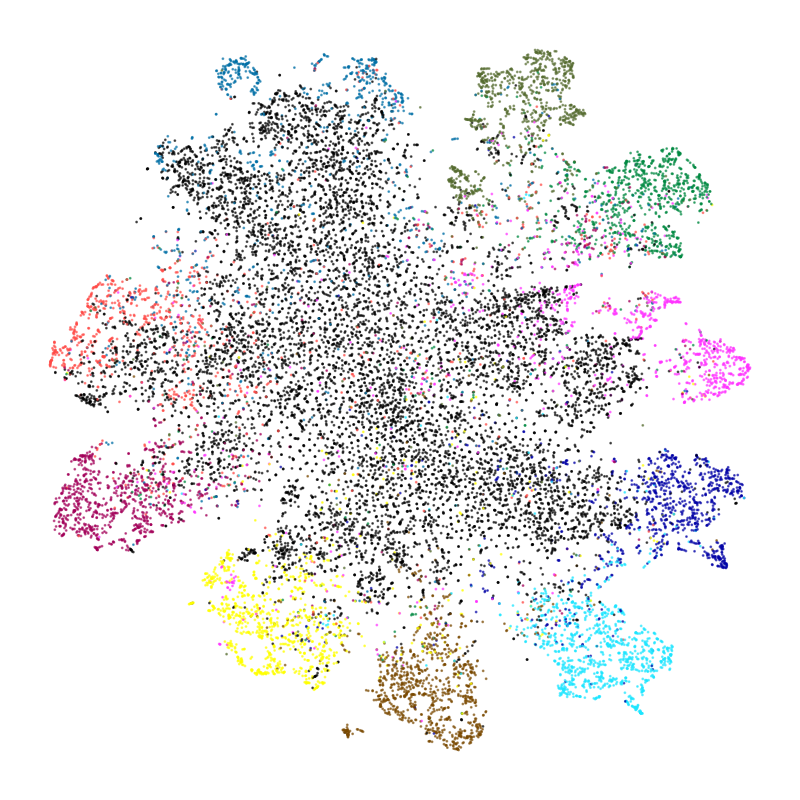}}
    \subfloat[Ours]{\includegraphics[width=0.49\linewidth]{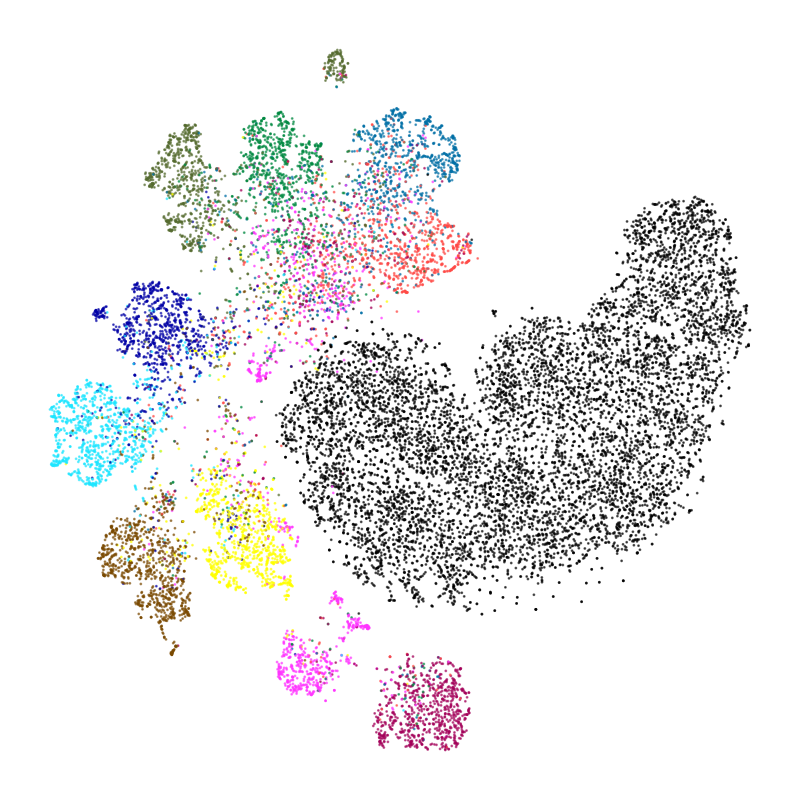}}
    \caption{T-SNE visualization on Cifar10-C test set with SVHN as strong OOD samples. Black dots are strong OOD examples while other colors are weak OOD examples. The colors indicate different classes.}
    \label{fig:TSNE}
    \vspace{-0.4cm}
\end{figure}

\noindent\textbf{Comparison with fixed thresholds}:
We use different fixed thresholds $\tau_{f}$ to divide the visible and invisible classes and compare it with the performance of our adaptive threshold $\tau^*$. As shown in Fig.~\ref{fig:comparethreshold}, it is impossible to find a fixed threshold that outperforms the adaptive threshold for strong OOD detection.

\noindent\textbf{T-SNE Visualization} To demonstrate the representation adaptation effect of different methods, we use t-SNE\cite{van2008visualizing} to reduce the dimensionality of the features for visualization. In Fig.~\ref{fig:TSNE}, we compared the features adapted by TEST~(without adaptation), TTAC \cite{su2022revisiting}, and our method. We observe a more distinct separation between weak~(colorful points) and strong~(black points) OOD sample features learned by our method, compared to without adaptation and state-of-the-art TTT method.
We also observe that filtering out strong OOD samples~(w/ OOD Det.) benefits the generic TTT method, e.g. (b) TTAC w/ OOD Det yields more distinguishable separation between weak and strong OOD samples of weak OOD and strong OOD features.

\begin{table}[!t]

    \caption{Performance under mixed strong OOD samples.}\label{tab:multiple}
    \vspace{-0.3cm}
    \scriptsize\centering{
    \resizebox{0.99\linewidth}{!}{
    \begin{tabular}{lccccccc}
    \toprule
    Metrics& TEST & BN & TENT & SHOT & TTT++ & TTAC & Ours \\
    \midrule
    $Acc_S$ & 63.90 & 75.18 & 58.10 & 76.63 & 60.58 & \textbf{85.01} & 82.19 \\
    $Acc_N$ & 88.20 & 90.37 & 77.19 & 60.94 & 87.39 & 66.78 & \textbf{96.76} \\
    $Acc_H$ & 74.11 & 82.08 & 66.30 & 67.89 & 71.56 & 74.80 & \textbf{88.88} \\
    \bottomrule
    \end{tabular}}}
    \vspace{-0.3cm}
\end{table}

\noindent\textbf{Performance Under Multiple Corruption Datasets}

We provide an evaluation, in Tab.~\ref{tab:multiple}, of combining ``snow'', ``contrast'', and ``glass blur'' on CIFAR10-C as weak OOD and combining random noise, MNIST, and SVHN as strong OOD.
It is evident that our approach consistently outperforms all competitors under both mixed weak OOD samples and mixed strong OOD samples.

\begin{figure}[!t]
    % \vspace{-0.3cm}
    \centering
    \includegraphics[width=0.99\linewidth]{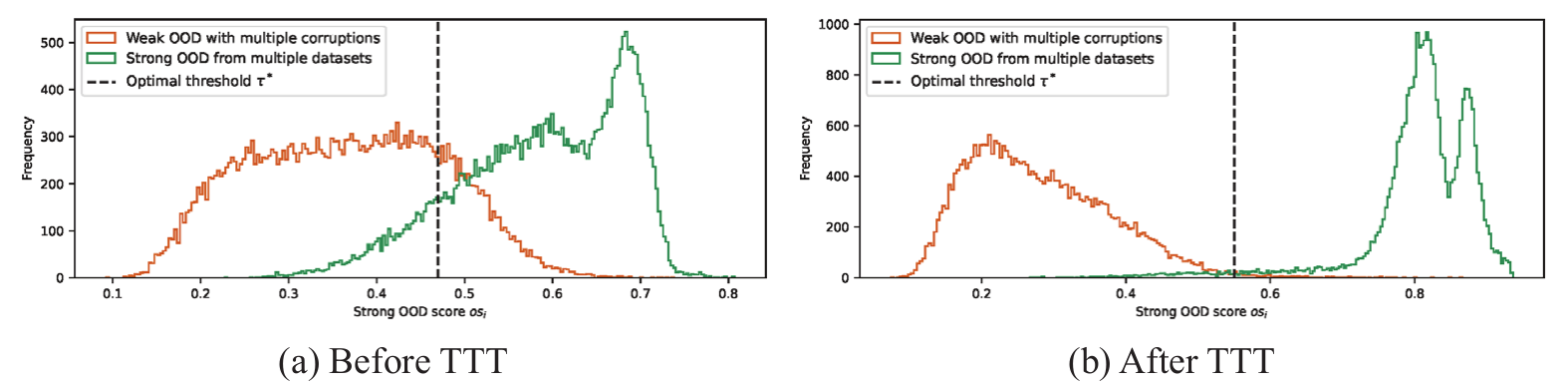}

    \vspace{-0.2cm}
    \caption{Noisy bimodal distribution under multiple corruption.}
    \label{fig:multiple_corruption}
    \vspace{-0.4cm}
\end{figure}

We illustrate the OOD score distributions in Fig.~\ref{fig:multiple_corruption} for the experiment with mixed weak OOD samples and mixed strong OOD samples presented in Tab.~\ref{tab:multiple}. In Fig.~\ref{fig:multiple_corruption}~(a) and (b) we respectively show the OOD score distributions before and after test-time training. We conclude that, although the more challenging mixed setting makes it harder to differentiate weak and strong OOD samples before TTT, our method can well separate these two groups after TTT. This suggests the robustness of our method under noisy OOD score distributions.

\section{Conclusion}
Test-time training has been extensively studied to enable adaptation to unknown target distribution with low inference latency. In this work, we investigate the robustness of test-time training when testing data is contaminated with strong OOD samples, a.k.a. open-world test-time training~(OWTTT). We proposed a hyper-parameter-free strong OOD detector that benefits both self-training and distribution alignment for OWTTT. We further allow the prototype pool to expand dynamically with which self-training can allow better separation between weak and strong OOD samples. Extensive evaluations on five OWTTT benchmarks demonstrated the effectiveness of the proposed method. 

\vspace{0.3cm}

\noindent\textbf{Acknowledgments.} This work is supported in part by the National
Natural Science Foundation of China (NSFC) under Grant 62106078, the A*STAR under Project C210112059, Sichuan Science and Technology Program under Project 2023NSFSC1421, Program for Guangdong Introducing Innovative and Entrepreneurial Teams (No.: 2017ZT07X183) and Guangdong R\&D key project of China (No.: 2019B010155001).

{\small
\bibliographystyle{ieee_fullname}
\bibliography{mainbib}
}

\newpage

\appendix

\setcounter{section}{0} 
\renewcommand{\thesection}{\Alph{section}}
\part*{Appendix}
\addcontentsline{toc}{part}{Appendix}

In this supplementary material, we first declare the limitations and failure cases for the proposed method. Then We perform more extensive experiments, including compatibility with the transformer-based model, ViT\cite{dosovitskiyimage}, robustness to different data streaming orders, and the evaluation results on 3D point cloud data. In addition, we further conduct some visualization experiments to provide a more intuitive demonstration of how our approach works.

\section{Limitation and Failure Case}
We explicitly discuss the limitation of the proposed method and analyze the failure case when the assumptions are violated. Our method is built to resolve a common scenario during open-world test-time training, i.e. target domain consists of strong OOD samples. Nevertheless, we do not rule out the scenario where the target domain is not contaminated with strong OOD samples. As our method relies on detecting strong OOD samples to improve robustness, it may mistakenly treat some weak OOD samples as strong OOD ones. As such, \textit{test-time training performance may be compromised when the target domain is not contaminated with strong OOD samples, and this is not a known priori}. To verify this limitation, we first evaluate our method, as well as other competing methods, on the ``clean'' CIFAR10-C test set, i.e. the target domain only contains the CIFAR10-C test set. To enable OOD detection under a potentially single-modal OOD score distribution, we restrict the strong OOD detection threshold to between $0.4$ and $1.0$. We compare the results with the target domain contaminated with random noise as strong OOD samples. As shown in Tab.~\ref{tab:clean}, we make the following observations from the results. First, when it is known a priori that the target domain only contains weak OOD samples, our method without using an OOD detector performs comparably to the state-of-the-art TTT methods, e.g. TTAC. However, all competing methods with strong OOD detectors would suffer when this prior knowledge is not available, e.g. Ours drops from $85.25\%$ to $78.63\%$ on clean accuracy. Nevertheless, our proposed method still outperforms all competing methods with a large margin on the harmonic mean $Acc_H$ and there is a good balance between clean accuracy and accuracy under strong OOD samples when OOD detection is included. In contrast, without strong OOD detection, all methods fail to identify strong OOD samples. As direct calculating $Acc_H$ under this circumstance yields $Acc_H=0$, we use N/A to indicate this situation. Overall, it still remains an open question of how to trade off a balance between being robust to strong OOD samples and maintaining good performance when the target domain only contains clean OOD samples.

\begin{table}
\caption{The performance under CIFAR10-C with~(w/~Str.~OOD) and without~(w/o~Str.~OOD) strong OOD samples.}\label{tab:clean}
\scriptsize\centering{
\begin{tabular}{c  ccc }
\toprule
Method& OOD Det. & w/o~Str.~OOD~$Acc_S$ & w/~Str.~OOD~$Acc_H$\\
\midrule
TEST & $\times$ &70.59 &  N/A \\
BN & $\times$&79.73  & N/A\\
TENT &$\times$& 80.91 &  N/A\\
SHOT & $\times$&82.58  & N/A\\
TTT++ &$\times$ &80.74 & N/A\\
TTAC &$\times$ &\textbf{86.36} & N/A \\
Ours &$\times$ &\underline{85.25} & N/A  \\
\midrule
TEST &\checkmark&54.19 & 81.36\\
BN &\checkmark&74.22 & \underline{85.11}\\
TENT &\checkmark&76.70 & 32.77\\
SHOT &\checkmark&74.64 & 67.23\\
TTT++ &\checkmark&76.86 & 47.86\\
TTAC & \checkmark&\underline{78.45} & 70.35\\
Ours &\checkmark&\textbf{78.63} & \textbf{91.56}\\
\bottomrule
\end{tabular}}\end{table}

\section{Compatibility with Transformer Backbone}
In this section, we perform additional experiments with ViT backbone~\cite{dosovitskiyimage}. The ViT model pre-trained in the clean CIFAR-10 dataset is utilized as the source domain model. Then we test it on the Cifar10-C test set under the strongest corruption level. All experiments were conducted under our OWTTT protocol, where random noise, MNIST, SVHN, CIFAR100-C, and Tiny-ImageNet are respectively selected as strong OOD data. The results presented in Tab.~\ref{tab:vit} demonstrated that our method is compatible with a more advanced backbone network.

\begin{table*}
\caption{Open-world test-time training under ViT backbone.}\label{tab:vit}
\scriptsize\centering{
\resizebox{0.99\linewidth}{!}{
\begin{tabular}{c  ccc  ccc  ccc  ccc  ccc }
\toprule
\multirow{2}*{Method}& \multicolumn{3}{c}{Noise}& \multicolumn{3}{c}{MNIST}& \multicolumn{3}{c}{SVHN}& \multicolumn{3}{c}{Tiny-ImageNet}& \multicolumn{3}{c}{CIFAR100-C}\\
\cmidrule{2-16}
&$Acc_S$&$Acc_N$&$Acc_H$&$Acc_S$&$Acc_N$&$Acc_H$&$Acc_S$&$Acc_N$&$Acc_H$&$Acc_S$&$Acc_N$&$Acc_H$&$Acc_S$&$Acc_N$&$Acc_H$\\
\cmidrule(lr){2-4}\cmidrule(lr){5-7}\cmidrule(lr){8-10}\cmidrule(lr){11-13}\cmidrule(lr){14-16}

TEST&86.22&\textbf{100.00}&92.60&82.24&\textbf{96.52}&\underline{88.81}&78.82&\underline{92.71}&85.20&82.29&\underline{71.41}&\underline{76.46}&80.48&\underline{76.50}&78.44\\
TENT&88.56&99.93&93.90&\textbf{87.49}&\underline{91.65}&\textbf{89.52}&75.98&49.51&59.95&84.16&61.63&71.16&78.91&56.50&65.85\\
SHOT&89.37&85.88&87.59&\underline{85.68}&76.18&80.65&78.17&50.93&61.67&\underline{89.22}&63.13&73.94&\textbf{86.44}&62.96&72.85\\
TTAC&\underline{90.14}&\textbf{100.00}&\underline{94.81}&77.28&54.01&63.58&\underline{85.03}&92.51&\underline{88.61}&85.55&68.09&75.82&\underline{85.30}&74.06&\underline{79.29}\\
OURS&\textbf{92.47}&\textbf{100.00}&\textbf{96.09}&73.67&65.22&69.19&\textbf{89.53}&\textbf{98.50}&\textbf{93.80}&\textbf{90.30}&\textbf{78.75}&\textbf{84.13}&83.12&\textbf{82.97}&\textbf{83.05}\\

\bottomrule
\end{tabular}}}\end{table*}

\begin{table*}[h!]
\caption{The performance of our method under different random seeds.}\label{tab:seed}
\scriptsize\centering{
\begin{tabular}{c  ccc  ccc  ccc  ccc  ccc }
\toprule
\multirow{2}*{Seed}& \multicolumn{3}{c}{Noise}& \multicolumn{3}{c}{MNIST}& \multicolumn{3}{c}{SVHN}& \multicolumn{3}{c}{Tiny-ImageNet}& \multicolumn{3}{c}{CIFAR100-C}\\
\cmidrule{2-16}
&$Acc_S$&$Acc_N$&$Acc_H$&$Acc_S$&$Acc_N$&$Acc_H$&$Acc_S$&$Acc_N$&$Acc_H$&$Acc_S$&$Acc_N$&$Acc_H$&$Acc_S$&$Acc_N$&$Acc_H$\\
\cmidrule(lr){2-4}\cmidrule(lr){5-7}\cmidrule(lr){8-10}\cmidrule(lr){11-13}\cmidrule(lr){14-16}
\#1&85.46&98.60&91.56&83.89&97.83&90.32&84.99&87.94&86.44&71.77&84.71&77.70&74.08&84.64&79.01\\
% 10&84.48&97.75&90.63&83.60&98.74&90.54&85.11&85.71&85.41&72.59&82.41&77.19&67.55&85.81&75.60\\
\#2&85.00&98.40&91.21&84.40&99.12&91.17&85.19&88.38&86.76&72.63&83.25&77.58&75.69&85.09&80.11\\
\#3&85.57&98.79&91.71&84.48&99.01&91.17&85.26&87.94&86.58&72.46&82.37&77.10&73.89&84.09&78.66\\
\#4&85.35&98.37&91.40&84.04&98.14&90.54&85.29&89.62&87.40&71.82&84.09&77.47&75.00&85.74&80.01\\
\bottomrule
\end{tabular}}\end{table*}

\begin{table*}[h!]
\caption{The performance of our method under different thresholding rates.}\label{tab:thresholding}
\scriptsize\centering{
\begin{tabular}{c  ccc  ccc  ccc  ccc  ccc }
\toprule
\multirow{2}*{Rate}& \multicolumn{3}{c}{Noise}& \multicolumn{3}{c}{MNIST}& \multicolumn{3}{c}{SVHN}& \multicolumn{3}{c}{Tiny-ImageNet}& \multicolumn{3}{c}{CIFAR100-C}\\
\cmidrule{2-16}
&$Acc_S$&$Acc_N$&$Acc_H$&$Acc_S$&$Acc_N$&$Acc_H$&$Acc_S$&$Acc_N$&$Acc_H$&$Acc_S$&$Acc_N$&$Acc_H$&$Acc_S$&$Acc_N$&$Acc_H$\\
\cmidrule(lr){2-4}\cmidrule(lr){5-7}\cmidrule(lr){8-10}\cmidrule(lr){11-13}\cmidrule(lr){14-16}
25\%&84.25&97.64&90.45&83.68&97.65&90.13&84.88&84.88&84.88&72.83&80.04&76.26&75.0&79.83&77.34\\
50\%&85.39&98.69&91.56&83.89&97.71&90.27&85.00&88.03&86.49&71.77&84.71&77.70&74.22&84.39&78.98\\
75\%&85.87&98.82&91.89&84.22&97.95&90.57&85.00&90.99&87.89&69.44&84.94&76.41&72.69&86.70&79.08\\
100\%&85.17&97.12&90.76&84.07&97.63&90.35&84.56&92.84&88.51&67.25&85.37&75.23&69.12&87.50&77.23\\

\bottomrule
\end{tabular}}\end{table*}

\section{Data Streaming Order}
In this section, we explore the impact of the testing data streaming order on our approach. We randomly shuffled the test data four times and performed test-time Training separately. The experimental results are shown in Tab.~\ref{tab:seed}. Since we use a moving average queue ${N_m}$ to select the optimal threshold $\tau^*$, which is less affected by the data streaming order, our method demonstrates strong stability regardless of the order of data streaming.

\section{Imapct of Thresholding Ratio}
To further reduce the effect of incorrect pseudo labeling, we only use 50$\%$ of samples with $od_i$ far from $\tau^*$ to perform prototype clustering for each batch. We explored the proportions of testing samples  for clustering by setting the proportions to 25\% 50\% 75\% and 100\%, and using CIFAR10-C as the weak OOD. The results are presented in Tab.~\ref{tab:thresholding}. It is evident that our method is not sensitive to  the proportion of used pseudo labels.

\section{Additional Details}

\noindent\textbf{Source Domain Prototypes}: We obtain the source domain prototypes by first running inference on all source domain training samples. More specifically, the prototypes are obtained via the following equation.

\begin{equation}
    p_k=\frac{1}{\sum\limits_{y_i\in\set{D}_s}\mathbbm{1}(y_i=k)}\sum\limits_{x_i,y_i\in\set{D}_s} \mathbbm{1}(y_i=k)\cdot f(x_i)
\end{equation}

\begin{table}
\caption{Open-world test-time training on point cloud data.}\label{tab:point_cloud}
\scriptsize\centering{
\begin{tabular}{l  cccccc }
\toprule
\multirow{2}*{Method}& \multicolumn{3}{c}{Noise}& \multicolumn{3}{c}{3DMNIST}\\
\cmidrule{2-4}\cmidrule(lr){5-7}
&$Acc_S$&$Acc_N$&$Acc_H$&$Acc_S$&$Acc_N$&$Acc_H$\\
\cmidrule{2-7}
TEST&44.77&88.77&59.52&42.11&\underline{79.94}&55.16\\
BN&58.06&85.20&69.06&47.81&69.72&56.73\\
TENT&19.50&60.37&29.47&17.33&57.22&26.60\\
SHOT&\underline{62.75}&79.79&\underline{70.25}&\textbf{61.58}&78.74&\underline{69.11}\\
TTAC&49.99&\underline{87.20}&63.55&43.97&77.02&55.98\\
OURS&\textbf{69.15}&\textbf{93.39}&\textbf{79.46}&\underline{59.24}&\textbf{88.95}&\textbf{71.12}\\
\bottomrule
\end{tabular}}\end{table}

\section{Evaluation on 3D Point Cloud Data}

To demonstrate the applicability of our proposed method across diverse tasks, we extended the OWTTT protocol to the 3D point cloud classification task.

\noindent\textbf{Method}: We maintain consistency with the methods in the manuscript, making only minor adaptations specifically for 3D point cloud data. Due to the inherently discrete nature of point cloud data compared to image data, we employ strong OOD prototypes $\set{P}_u $ and weak OOD prototypes $\set{P}_s$ together to enhance the discriminative power of the Strong OOD Score $OS'_i$, defined as Eq.~\ref{ood_score_new}. To enable the adaptation of $\set{P}_u$ with respect to changes in the test data, we employ a momentum-based updating approach. Specifically, for a given sample $x_s$ that is predicted as a strong OOD instance, we update the most similar strong OOD prototype $p_i$ in $\set{P}_u$, as shown in Eq.~\ref{momentum_update}.

\begin{equation}
\label{ood_score_new}
\begin{aligned}
    OS'_{i} &= (1 - s_{s}) \cdot \frac{s_{s}}{s_{s} + s_{u}} + s_{u} \cdot \frac{s_{u}}{s_{s} + s_{u}}\\
    s_s &= \max\limits_{p_k\in\set{P}_s} <f(x_i),p_k> \\
    s_u & = \frac{1}{10}\sum_{j=1}^{10} d_{ij} \\
    & s.t.\quad \{d_{ij}\}_{j=0}^{|\set{P}_u|-1} = sort(\{<f(x_i), p_k>\}_{p_k \in \set{P}_u}) \\
    & and\quad d_{i0} = \max\limits_{p_k\in\set{P}_u} <f(x_i), p_k> \\
\end{aligned}
\end{equation}

\begin{equation}
\label{momentum_update}
    p_i = (1 - \delta)p_i + \delta f(x_s)
\end{equation}

\noindent\textbf{Datasets}: We choose ModelNet40-C \cite{sun2021benchmarking} as weak OOD, which consists of 15 common corruptions of point cloud data, with 9,843 training samples and 2,468 test samples. We select random noise and the 3D representation of MNIST \cite{3DMNIST} as strong OOD.

\noindent\textbf{Training Details}: We follow \cite{su2022revisiting} and use the DGCNN \cite{wang2019dynamic}, with learning rate $\alpha$=1e-4, batch size $N_B$ = 64, $\lambda$ = 1.

\noindent \textbf{Results}: We observe from the results in Tab.~\ref{tab:point_cloud} that our method outperforms all competing methods on the point cloud datasets. The results demonstrate that our proposed method also exhibits a strong fit for 3D point cloud data, showcasing its potential for broader application in various fields.

\section{Adaptive Threshold VS Fixed Threshold} 
\begin{figure}[!htb]
    \centering
    % \vspace{-0.3cm}
    \includegraphics[width=0.99\linewidth]{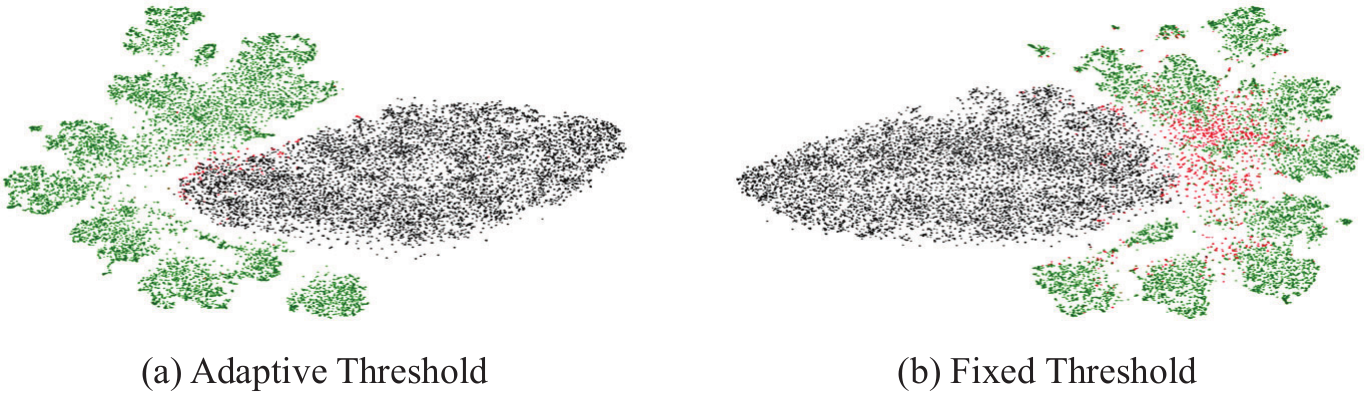}
    % \vspace{-0.4cm}
    \caption{T-SNE visualizations of adaptive threshold and fixed threshold on CIFAR10-C with SVHN as strong OOD.}
    \label{fig:adaptive}
    % \vspace{-0.4cm}
\end{figure}

We visualize testing samples on CIFAR10-C with SVHN as strong OOD via t-SNE in Fig.~\ref{fig:adaptive} to compare the fixed threshold and our adaptive threshold. \textcolor[rgb]{0,0.4,0}{Green}, black and \textcolor[rgb]{0.8,0,0}{red} dots indicate correctly classified weak OOD samples, correctly classified strong ODD samples and misclassified samples respectively. We clearly observe fewer misclassified samples with adaptive thresholds, suggesting the advantage.

\section{Dynamic Representations} 
\begin{figure}[!htb]
    \centering

    \includegraphics[width=0.99\linewidth]{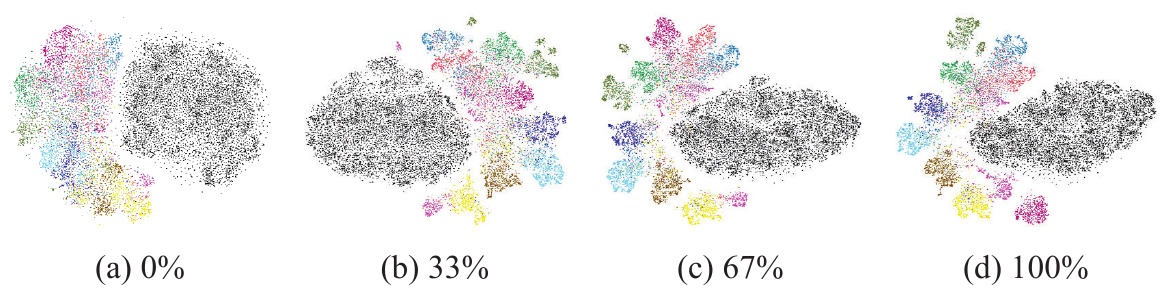}

    \caption{T-SNE visualizations on CIFAR10-C with SVHN as strong OOD samples as TTT progresses.}
    \label{fig:cumulative}

\end{figure}

We further present a t-SNE visualization at four different stages (indicated by the percentage TTT progress) of TTT in Fig.~\ref{fig:cumulative}. It is obvious that different semantic classes (colorful dots) become better separated as TTT progresses and the strong OOD samples (black dots) are always well separated from weak OOD ones.

\end{document}